\documentclass{article}
\usepackage{microtype}
\usepackage{graphicx}
\usepackage{subfigure}
\usepackage{booktabs} % for professional tables                                                                                                   
\usepackage{hyperref}

\usepackage{latexsym}
\usepackage{url}
\usepackage{amssymb}
\usepackage{amsmath,amsthm}
\usepackage{bm}

\usepackage{appendix}
\usepackage{natbib}
\newcommand{\cutthepeakeroff}[1]{}
\newcommand{\source}{\mathbf{x}}
\newcommand{\sourcek}{x}
\newcommand{\target}{\mathbf{t}}
\newcommand{\targetk}{t}
\newcommand{\hidenc}{\mathbf{z}}
\newcommand{\hidenck}{z}
\newcommand{\hiddec}{\mathbf{h}}
\newcommand{\hiddeck}{h}

\usepackage[accepted]{icml2018}                                                                                                                                                  

\icmltitlerunning{Analyzing Uncertainty in Neural Machine Translation}

\begin{document}

\twocolumn[
\icmltitle{Analyzing Uncertainty in Neural Machine Translation}

\icmlsetsymbol{equal}{*}

\begin{icmlauthorlist}
\icmlauthor{Myle Ott}{fb}
\icmlauthor{Michael Auli}{fb}
\icmlauthor{David Grangier}{fb}
\icmlauthor{Marc'Aurelio Ranzato}{fb}
\end{icmlauthorlist}
\icmlaffiliation{fb}{Facebook AI Research, USA}
\icmlcorrespondingauthor{Myle Ott}{myleott@fb.com}

\icmlkeywords{machine translation, uncertainty}

\vskip 0.3in
]



\printAffiliationsAndNotice{}  % leave blank if no need to mention equal contribution                                                                                            
\begin{abstract}

% Machine translation has become a popular task in machine learning for its use as a test bed for neural sequence-to-sequence models. Despite abundant recent work, there is a lack of understanding about how these models work. Practitioners have reported several unexplained observations, including the performance degradation of large beams, the under-estimation of rare words and the lack of diversity of generated hypotheses.

% In this work, we demonstrate that many of these findings are related to the inherent uncertainty of the task, due to the existence of multiple valid translations for a single source sentence, and to extrinsic uncertainty due to noisy training data of modern benchmarks. We propose tools and metrics to assess how uncertainty affects the model distribution and search strategies
% used to generate translations. We show that while search works remarkably well, NMT models tend to spread too much probability mass in the hypothesis space.

% We propose tools to assess model calibration and show how to easily fix some shortcomings of current models. We release
% both code and multiple human reference translations for two popular benchmarks.
\textit{%
Machine translation is a popular test bed for research in neural sequence-to-sequence models but despite much recent research, there is still a lack of understanding of these models.
Practitioners report performance degradation with large beams, the under-estimation of rare words and a lack of diversity in the final translations.
Our study relates some of these issues to the inherent uncertainty of the task, due to the existence of multiple valid translations for a single source sentence, and to the extrinsic uncertainty caused by noisy training data.
We propose tools and metrics to assess how uncertainty in the data is captured by the model distribution and how it affects search strategies that generate translations.
Our results show that search works remarkably well but that models tend to spread too much probability mass over the hypothesis space.
Next, we propose tools to assess model calibration and show how to easily fix some shortcomings of current models.
%We release both code and multiple human reference translations for two popular benchmarks.
As part of this study, we release multiple human reference translations for two popular benchmarks. 
}
\end{abstract}

\section{Introduction} \label{sec:introduction}
% MT is an interesting ML task
Machine translation (MT) is an interesting task not only for its practical applications but also for the formidable learning challenges it poses, from how to transduce
variable length sequences, to searching for likely sequences in an intractably large hypothesis space, to dealing with the multi-modal nature of the prediction task,
since typically there are several correct ways to translate a given sentence.

% Lots of work on the topic, what we do
The research community has made great advances on this task, recently focusing the effort on the exploration of several variants of neural models~\citep{bahdanau2014neural,luong2015effective,gehring2017convs2s,vaswani2017transformer} that have greatly improved the state of the art performance on public benchmarks.
However, several open questions remain~\citep{koehn17}. In this work, we analyze top-performing trained models in order to answer some of these open questions. We target better understanding to help prioritize future exploration towards important aspects of the problem and therefore speed up progress.

% Known open issues and questions they raise which we answer
For instance, according to conventional wisdom neural machine translation (NMT) systems under-estimate rare words~\citep{koehn17}, why is that?
Is the model poorly calibrated? Is this due to exposure bias~\citep{mixer}, i.e., the mismatch between the distribution of words observed at training and test time?
Or is this due to the combination of uncertainty in the prediction of the next word and inference being an $\arg\max$ selection process, which always picks the most likely/frequent word?
Similarly, it has been observed~\citep{koehn17} that performance degrades with large beams.
Is this due to poor fitting of the model which assigns large probability mass to bad sequences? Or is this due to the heuristic nature of this search procedure which
fails to work for large beam values?
In this paper we will provide answers and solutions to these and other related questions.

% Uncertainty is key: not only generalization but also dealing with an output space where multiple hyotheses have non negligibl probab mass.
% Which hypotheses is search going to find? How does the model fit such a distribution? How wll is the model calibrated?
% What tools to use? tough eneavor given we obsrve handful of references if not only one
The underlying theme of all these questions is \textit{uncertainty}, i.e. the one-to-many nature of the learning task.
In other words, for a given source sentence there are several target sequences that have non negligible probability mass. Since the model only observes one or very
few realizations from the data distribution, it is natural to ask the extent to which an NMT model trained with token-level cross-entropy is able to capture such a rich distribution,
and whether the model is calibrated. Also, it is equally important to understand the effect that uncertainty has on search and whether there are better and more efficient search strategies.

% Findings and contributions
% Develop suite of tools to ananlyze both search and model
% Find search to work remarkable well, much more effective and efficient than sampling; in fact, for the same logprob beam gets higher BLEU in average
% Find model to be overall remarkably well calibrated at the set level
% Model spreads probab too much, often resulting in under-estimation of individual hypothesis
% samples all over the place with low quality, beam better quality but lacks diversity
% extrinsic noise causes degradation of large beam, and it can be easily fixed
Unfortunately, NMT models have hundreds of millions of parameters, the search space is exponentially large and we typically observe only one reference for a given
source sentence. Therefore, measuring fitness of a NMT model to the data distribution is a challenging scientific endeavor, which we tackle by borrowing and combining
tools from the machine learning and statistics literature~\citep{structpredcalib, guo17}. With these tools, we show that search works surprisingly well, yielding highly likely sequences even with relatively narrow beams. Even if we consider samples from the model that have similar likelihood, beam hypotheses yield higher BLEU on average.
Our analysis also demonstrates that although NMT is well calibrated at the token and set level, it generally spreads too much probability mass over the space of sequences. This often results in individual hypotheses being under-estimated, and overall, poor quality of samples drawn from the model.
Interestingly, systematic mistakes in the data collection process also contribute to uncertainty, and
a particular such kind of noise, the target sentence being replaced by a copy of the corresponding source sentence, 
is responsible for much of the degradation observed when using wide beams.

% Conclusion: we can analyze despite having access to only one or few references,
% search is very effective in current NMT but we need to concenrate probab mass, watch out for effects of noise in the data collection process,
This analysis -- the first one of its kind -- introduces tools and metrics to assess fitting of the model to the data distribution, and shows areas of improvement for NMT. It also suggests easy fixes for some of the issues reported by practitioners. We also release the data we collected for our evaluation, which consists of ten human translations for 500 sentences taken from the WMT'14 En-Fr and En-De test sets.\footnote{Additional reference translations are available from:\\
\url{https://github.com/facebookresearch/analyzing-uncertainty-nmt}.}

\section{Related Work}
In their seminal work, \citet{zoph15} frame translation as a compression game and measure the amount of information added by translators. While this work precisely quantifies
the amount of uncertainty, it does not investigate its effect on modeling and search. In another context, uncertainty has been considered for the design of better evaluation
metrics~\citep{hyter,galley2015deltableu}, in order not to penalize a model for producing a valid translation which is different from the provided reference.

Most work in NMT has focused on improving accuracy without much consideration for the intrinsic uncertainty of the translation task itself~\citep{bahdanau2014neural,luong2015effective,gehring2017convs2s,vaswani2017transformer} (\textsection\ref{sec:data_unc}).
Notable exceptions are latent variable models~\citep{blunsom08,vae_nmt} which explicitly attempt to model multiple modes in the data distribution,
or decoding strategies which attempt to predict diverse outputs while leaving the model unchanged~\citep{gimpel13,diverse_bs_batra16,diverse_bs_li16,npad}.
However, none of these works check for improvements in the match between the model and the data distribution.

Recent work on analyzing machine translation has focused on topics such as comparing neural translation to phrase-based models~\citep{bentivogli16, toral17}.
\citet{koehn17} presented several challenges for NMT, including the deterioration of accuracy for large beam widths and the under-estimation of rare words, which we address
in this paper. \citet{isabelle17} propose a new evaluation benchmark to test whether models can capture important linguistic properties. %
Finally, \citet{niehues17} focus on search and argue in favor of better translation modeling instead of improving search.

\section{Data Uncertainty}
\label{sec:data_unc}

Uncertainty is a core challenge in translation, as there are several ways to correctly translate a sentence; but what are typical sources of uncertainty found in modern benchmark datasets?
Are they all due to different ways to paraphrase a sentence?
In the following sections, we answer these questions, distinguishing uncertainty inherent to the task itself (\textsection\ref{sec:intrinsic}), and uncertainty due to spurious artifacts caused by the data collection process (\textsection\ref{sec:extrinsic}).

\subsection{Intrinsic Uncertainty}
\label{sec:intrinsic}

One source of uncertainty is the existence of several \textit{semantically equivalent} translations of the same source sentence.
This has been extensively studied in the literature~\citep{hyter,pado09}.
Translations can be more or less literal, and even if literal there are many ways to express the same meaning.
Sentences can be in the active or passive form and for some languages determiners and prepositions such as `the', `of', or `their' can be optional.

Besides uncertainty due to the existence of distinct, yet semantically equivalent translations, there are also sources of uncertainty due to \textit{under-specification} when translating
into a target language more inflected than the source language.
Without additional context, it is often impossible to predict the missing gender, tense, or number, and therefore, there are multiple plausible translations of the same source sentence. Simplification or addition of cultural context are also common sources of uncertainty~\citep{venuti08}.

\subsection{Extrinsic Uncertainty}
\label{sec:extrinsic}

Statistical machine translation systems, and in particular NMT models, require lots of training data to perform well.
To save time and effort, it is common to augment high quality human translated corpora with lower quality web crawled data~\citep{smith2013dirt}.
This process is error prone and responsible for introducing additional uncertainty in the data distribution.
Target sentences may only be partial translations of the source, or the target may contain information not present in the source.
A lesser-known example are target sentences which are entirely in the source language, or which are primarily \textit{copies of the corresponding source}.
For instance, we found that between 1.1\% to 2.0\% of \textit{training} examples in the WMT'14 En-De and WMT'14 En-Fr datasets (\textsection\ref{sec:datasets}) are ``copies'' of the source sentences, where a target sentence is labeled as ``copy'' if the intersection over the union of unigrams (excluding punctuation and numbers) is at least 50\%.
Source copying is particularly interesting since we show that, even in small quantities, it can significantly affect the model output (\textsection\ref{sec:beam_failure}). Note that \textit{test} sets are manually curated and never contain copies.

\section{Experimental Setup}
\label{sec:setting}

\subsection{Sequence to Sequence Model}
\label{sec:modeldesc}
Our experiments rely on the pre-trained models of the \emph{fairseq-py} 
toolkit~\citep{gehring2017convs2s}, which achieve competitive performance on the datasets we consider.
Formally, let $\source$ be an input sentence with $m$ words $\{\sourcek_1, \dots, \sourcek_m\}$,
and $\target$ be the ground truth target sentence with $n$ words $\{\targetk_1, \dots, \targetk_n\}$.
The model is composed of an \textit{encoder} and a \textit{decoder}.
The encoder takes $\source$ through several convolutional layers to produce a sequence of hidden states, $\hidenc = \{\hidenck_1, \dots, \hidenck_m\}$, one per input word.
At time step $k$, the decoder takes a window of words produced so far (or the ground truth words at training time),
 $\{\targetk_{k-1}, \dots, \targetk_{k-i}\}$, the set of encoder hidden states $\hidenc$ and produces a distribution over the current
word: $p(\targetk_{k} | \targetk_{k-1}, \dots, \targetk_{k-i}, \hidenc)$.
More precisely, at each time step, an \textit{attention} module~\citep{bahdanau2014neural} summarizes the sequence $\hidenc$ with a single vector through a weighted sum of $\{\hidenck_1, \dots, \hidenck_m\}$. The weights depend on the source sequence $\source$ and the decoder hidden state, $\hiddeck_{k}$, which is the output of several convolutional layers taking as input
$\{\targetk_{k-1}, \dots, \targetk_{k-i}\}$.
From the source attention vector, the hidden state of the decoder is computed and the model emits a distribution over the current word as in:
$p(\targetk_{k} | \hiddeck_{k}) = \mbox{softmax}(W \hiddeck_{k} + b)$. \citet{gehring2017convs2s} provides further details.
To train the translation model, we minimize the cross-entropy loss:
$\mathcal{L} = -\sum_{i=1}^n \log p(\targetk_i | \targetk_{i-1}, \dots, \targetk_{1}, \source)$, using Nesterov's momentum~\citep{sutskever2013icml}.\footnote{We also obtain similar results with models trained with sequence-level losses~\cite{edunov2018classical}.}

At test time, we aim to output the most likely translation given the source sentence, according to the model estimate. We approximate such an output via \textit{beam search}.
Unless otherwise stated, we use beam width $k=5$, where hypotheses are selected based on their length-normalized log-likelihood.
Some experiments consider \emph{sampling} from the model conditional distribution $p(\targetk_{i}|\targetk_{i-1}, \hiddec_{i-1}, \source)$, one token at a time,
until the special end of sentence symbol is sampled.

\subsection{Datasets and Evaluation}
\label{sec:datasets}
We consider the following datasets:

\noindent {\bf WMT'14 English-German (En-De):} We use the same setup as~\citet{luong2015effective} which comprises 4.5M sentence pairs for training and we test on newstest2014.
We build a validation set by removing 44k random sentence-pairs from the training data.
As vocabulary we use 40k sub-word types based on a joint source and target byte pair encoding (BPE; Sennrich et al., 2016)\nocite{bpe}.

\noindent {\bf WMT'17 English-German (En-De):} The above pre-processed version of WMT'14 En-De did not provide a split into sub-corpora which we required for some experiments.
We therefore also experiment on the 2017 data where we test on newstest2017.
The full version of the dataset (\emph{original}) comprises 5.9M sentence pairs after length filtering to 175 tokens. We then consider the news-commentary portion with 270K sentences (\emph{clean}), and a \emph{filtered} version comprising 4M examples after removing low scoring sentence-pairs according to a model trained on the cleaner news-commentary portion.

\noindent  {\bf WMT'14 English-French (En-Fr):}
We remove sentences longer than 175 words and pairs with a source/target length ratio exceeding 1.5 resulting in 35.5M sentence pairs for training.
The source and target vocabulary is based on 40k BPE types. Results are reported on both newstest2014 and a validation set held-out from the training data comprising
26k sentence pairs.

We evaluate with tokenized BLEU~\citep{bleu} on the corpus-level and the sentence-level, after removing BPE splitting.
Sentence-level BLEU is computed similarly to corpus BLEU, but with smoothed $n$-gram counts (+1) for $n>1$~\citep{lin2004orange}.

\section{Uncertainty and Search} \label{sec:beam}
\begin{figure*}[t]
\centering
\vspace{-.25cm}
\includegraphics[width=\linewidth]{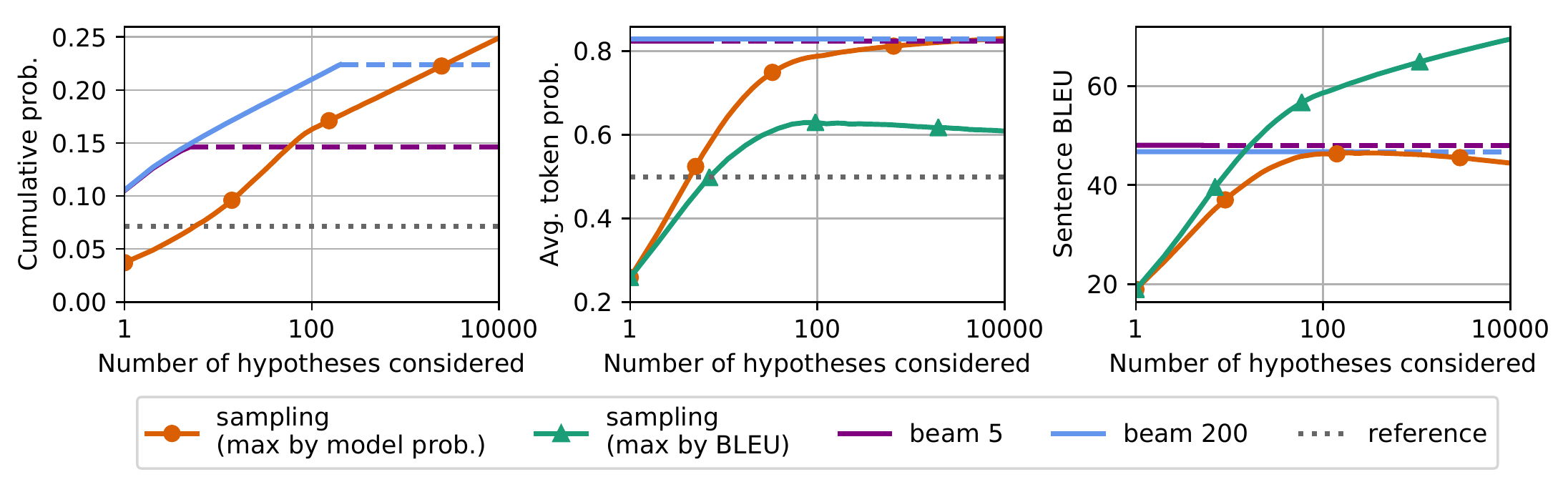}
\vspace{-1cm}
\caption{
\small
\emph{Left}: Cumulative sequence probability of hypotheses obtained by beam search and sampling on the WMT'14 En-Fr valid set;
\emph{Center}: same, but showing the average per-token probability as we increase the number of considered hypotheses, 
for each source sentence we select the hypothesis with the maximum probability (orange) or sentence-level BLEU (green); 
\emph{Right}: same, but showing averaged sentence-level BLEU as we increase the number of hypotheses.
}
\label{fig:oracle_bleu}
\end{figure*}
In this section we start by showing that the models under consideration are well trained (\textsection\ref{sec:welltrained}).
Next, we quantify the amount of uncertainty in the model's output and compare two search strategies: beam search and sampling (\textsection\ref{sec:beamisgood}).
Finally we investigate the influence of a particular kind of extrinsic uncertainty in the data on beam search, and provide an explanation 
for the performance degradation observed with wide beams (\textsection\ref{sec:beam_failure}).
\begin{table}[!t]%[!h]
\small
\centering
\begin{tabular}{lcc}
\toprule
& En-Fr & En-De \\
\midrule
\multicolumn{1}{l}{\bf Automatic evaluation} \\
\quad train PPL    & 2.54 & 5.14 \\
\quad valid PPL    & 2.56 & 6.36 \\
%\quad test BLEU    & 41.03 & 24.78 \\
\quad test BLEU    & 41.0 & 24.8 \\
\midrule
\multicolumn{2}{l}{\bf Human evaluation (pairwise)} \\
\quad Ref $>$ Sys  & 42.0\%  & 80.0\% \\
\quad Ref $=$ Sys  & 11.6\%  &  5.6\% \\
\quad Ref $<$ Sys  & 46.4\%  & 14.4\% \\
\bottomrule
\end{tabular}
\caption{\small Automatic and human evaluation on a 500 sentence subset of the WMT'14 En-Fr and En-De test sets.
Models generalize well in terms of perplexity and BLEU. Our human evaluation compares (reference, system) pairs for beam $5$.}
\label{tab:humanstudy}
\end{table}
\normalsize

\subsection{Preliminary: Models Are Well Trained}
\label{sec:welltrained}

We start our analysis by confirming that the models under consideration are well trained.
Table~\ref{tab:humanstudy} shows that the models, and particularly the En-Fr model, achieve low perplexity and high BLEU scores.

To further assess the quality of these models, we conducted a human evaluation with three professional translators.
Annotators were shown the source sentence, reference translation, and a translation produced by our model through \emph{beam search}---a breadth-first search that retains only the $k$ most likely candidates at each step.
Here, we consider a relatively narrow beam of size $k=5$.
The reference and model translations were shown in random order and annotators were blind to their identity.
We find that model translations roughly match human translations for the En-Fr dataset, while for the En-De dataset humans prefer the reference over the model output 80\% of the time.
Overall, the \textbf{models are well trained}---particularly the En-Fr model---and beam search can find outputs that are highly rated by human translators.

\subsection{Model Output Distribution Is Highly Uncertain}
\label{sec:beamisgood}

How much uncertainty is there in the model's output distribution?
What search strategies are most \emph{effective} (i.e., produce the highest scoring outputs) and \emph{efficient} (i.e., require generating the fewest candidates)?
To answer these questions we sample $10$k translations and compare them to those produced by beam search with $k=5$ and $k=200$.

Figure~\ref{fig:oracle_bleu} (\emph{Left}) shows that \textbf{the model's output distribution is highly uncertain}:
even after drawing $10$k samples we cover only 24.9\% of the sequence-level probability mass.
And while beam search is much more efficient at searching this space, covering 14.6\% of the output probability mass with $k=5$ and 22.4\% of the probability mass with $k=200$,
these finding suggest that most of the probability mass is spread elsewhere in the space (see also \textsection\ref{subsec:seq_level_calibration}).

Figure~\ref{fig:oracle_bleu} also compares the average sentence-level BLEU 
% (smoothed by adding 1 to every n-gram count) 
and model scores of hypotheses produced by sampling and beam search.
Sampling results for varying sample size $n=1, \dots, 10k$ are on two curves: orange reports probability (\emph{Center}) and sentence BLEU (\emph{Right}) for the sentence with the highest probability within $n$ samples, while green does the same for the sentence with the highest sentence BLEU in the same set~\citep{sokolov08}.
We find that sampling produces hypotheses with similar probabilities as beam search (\emph{Center}), however, for the same likelihood beam hypotheses have higher BLEU scores (\emph{Right}).
We also note that \textbf{BLEU and model probability are imperfectly correlated}:
while we find more likely translations as we sample more candidates, BLEU over those samples eventually decreases (\emph{Right}, orange curve).\footnote{Hypothesis length only decreases slightly with more samples, i.e., the BLEU brevity penalty moves from 0.975 after drawing 300 samples to 0.966 after 10k samples.}
Vice versa, hypotheses selected by BLEU have lower likelihood score beyond 80 samples (\emph{Center}, green curve).
We revisit this surprising finding in \textsection\ref{sec:beam_failure}.
%Accordingly, we restrict most of our analysis in later sections to hypotheses produced by beam search.

\begin{figure}[t]
\centering
\includegraphics[width=0.85\linewidth]{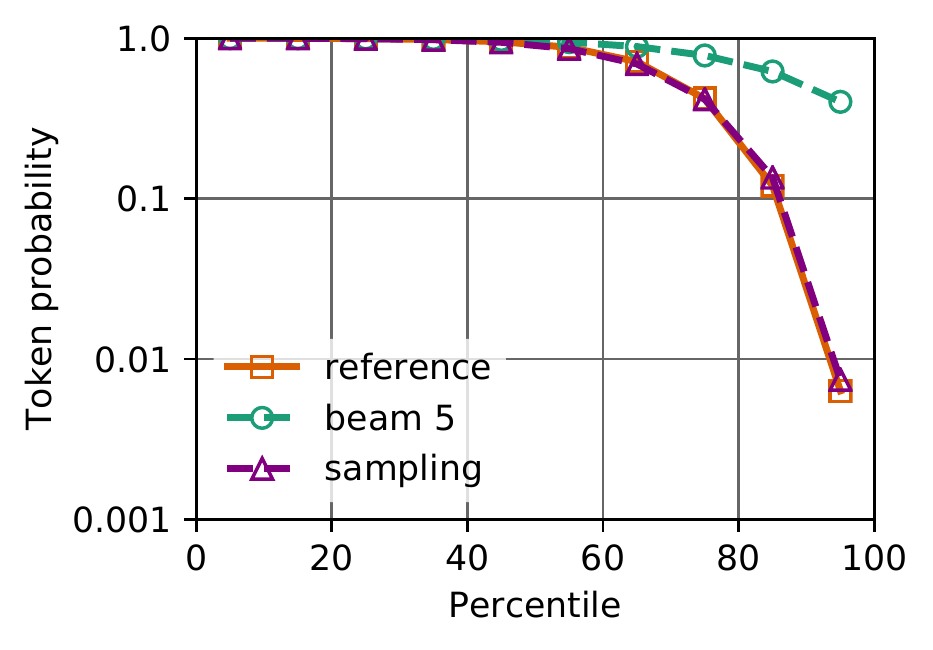}
\vspace{-0.5cm}
\caption{\small Probability quantiles for tokens in the reference, beam search hypotheses ($k=5$), and sampled hypotheses for the WMT'14 En-Fr validation set.
}
\vspace{-0.5cm}
\label{fig:logprob_quantile}
\end{figure}
Finally, we observe that the model on average assigns much lower scores to the reference translation compared to beam hypotheses (Figure~\ref{fig:oracle_bleu}, \emph{Center}).
% Finally, we observe that the model assigns much lower scores to the reference translation compared to either beam hypotheses or even just 5 samples, on average (Figure~\ref{fig:oracle_bleu}, \emph{Center}). % MR: not sure if it is really 5, and not sure this is consistent with fig.2.
%More surprisingly, sampling can produce hypotheses that score higher than the reference after just 5 samples on average.
%
To better understand this, in Figure~\ref{fig:logprob_quantile} we compare the token-level model probabilities of the reference translation, to those of outputs from beam search and sampling.
% at the token-level by plotting the sorted probabilities of all tokens in the WMT'14 En-Fr validation set.
%Figure~\ref{fig:logprob_quantile} shows that \textbf{beam search is a very effective search strategy}
We observe once again that \textbf{beam search is a very effective search strategy}, finding hypotheses with very high average token probabilities and rarely leaving high likelihood regions;
indeed only 20\% of beam tokens have probabilities below 0.7.
In contrast, the probabilities for sampling and the human references are much lower.
The high confidence of beam is somewhat surprising if we take into account the exposure bias~\citep{mixer} of these models, which have only seen gold translations at training time.
We refer the reader to \textsection\ref{subsec:seq_level_calibration} for discussion about how well the model actually fits the data distribution.
%Note however that even a small amount of uncertainty at the token-level can have a large effect at the sequence-level; e.g., a sentence with 30 words, each with probability 0.7, gets a sequence-level probability of only 2e-5.
%Indeed, 200 beam search hypotheses account for only 22.4\% of the sequence-level probability mass in average (Figure~\ref{fig:oracle_bleu}), suggesting that most of the \textbf{probability mass is spread} elsewhere in the space.

\subsection{Uncertainty Causes Large Beam Degradation}
\label{sec:beam_failure}
In the previous section we observed that repeated sampling from the model can have a negative impact on BLEU, even as we find increasingly likely hypotheses.
Similarly, we observe lower BLEU scores for beam 200 compared to beam 5, consistent with past observations about performance degradation with large beams~\citep{koehn17}.

Why does the BLEU accuracy of translations found by larger beams deteriorate rather than improve despite these sequences having higher likelihood?
To answer this question we return to the issue of extrinsic uncertainty in the training data (\textsection\ref{sec:extrinsic}) and its impact on the model and search. One particularly interesting case of noise is when target sentences in the training set are simply a copy of the source.

In the WMT'14 En-De and En-Fr dataset between 1.1\% and 2.0\% of the training sentence pairs are ``copies'' (\textsection\ref{sec:extrinsic}).
How does the model represent these training examples and does beam search find them? It turns out that \textbf{copies are over-represented in the output of beam search}.
On WMT'14 En-Fr, beam search outputs copies at the following rates: 2.6\% (beam=1), 2.9\% (beam=5), 3.2\% (beam=10) and 3.5\% (beam=20).

\begin{figure}[t]
\centering
\vspace{-.2cm}
\includegraphics[width=0.9\linewidth]{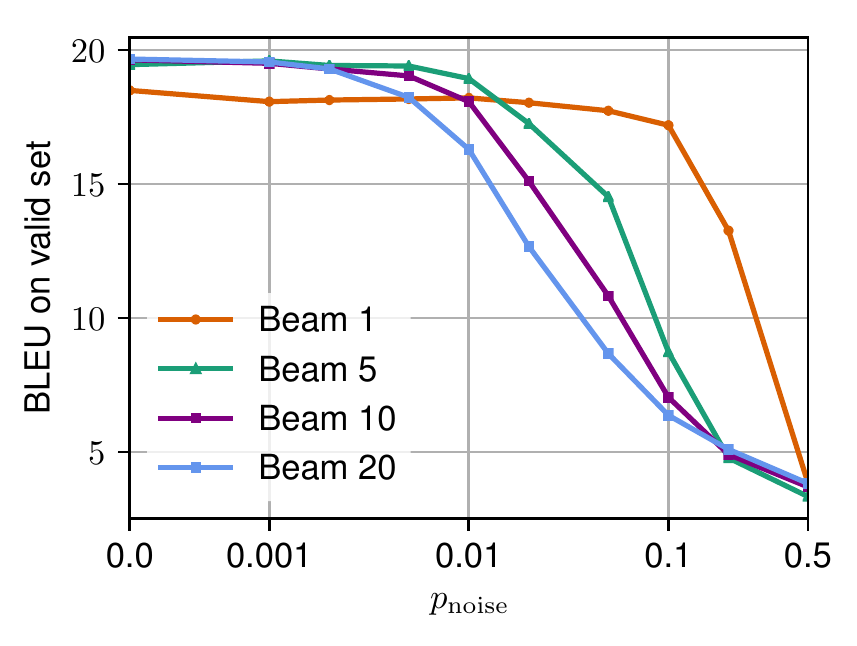}
\vspace{-.6cm}
\caption{\small Translation quality of models trained on WMT'17 English-German news-commentary data with added synthetic copy noise in the training data (x-axis) tested with various beam sizes on the validation set.}
\label{fig:copy_synth_beam}
\end{figure}

To better understand this issue, we trained models on the news-commentary portion of WMT'17 English-German which does not contain copies.
We added synthetic copy noise by randomly replacing the true target by a copy of the source with probability $p_{\rm noise}$. Figure~\ref{fig:copy_synth_beam} shows that larger beams are much more affected by copy noise.
Even just 1\% of copy noise can lead to a drop of 3.3 BLEU for a beam of $k=20$ compared to a model with no added noise. For a 10\% noise level, all but greedy search have their accuracy more than halved.

\begin{figure}[t]
\centering
\vspace{-.3cm}
\includegraphics[width=0.96\linewidth]{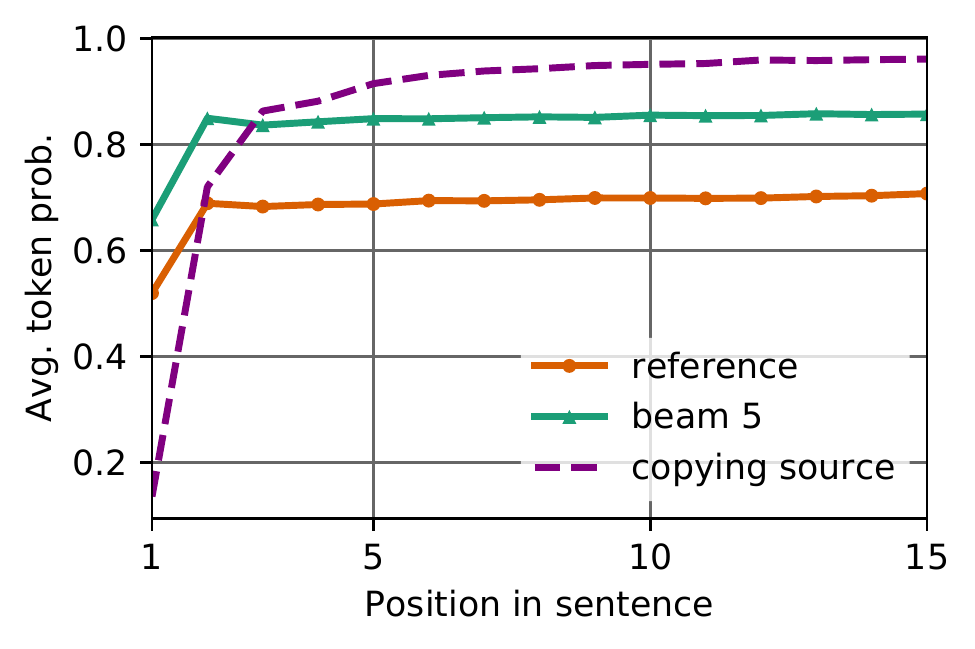}
\vspace{-.5cm}
\caption{\small Average probability at each position of the output sequence on the WMT'14 En-Fr validation set, comparing the reference translation, beam search hypothesis ($k=5$), and copying the source sentence.
}
%\vspace{-.25cm}
\label{fig:costlogprob}
\end{figure}

Next, we examine model probabilities at the token-level.
Specifically, we plot the average per position log-probability assigned by the En-Fr model to each token of: (i) the reference translation, (ii) the beam search output with $k=5$, and (iii) a synthetic output which is a copy of the source sentence.
Figure~\ref{fig:costlogprob} shows that the probability of copying the first source token is very unlikely according to the model (and actually matches the ground
truth rate of copy noise). However, after three tokens the model switches to almost deterministic transitions.
Because beam search proceeds in strict left-to-right manner, the copy mode is only reachable if the beam is wide enough to consider the first source word which has low probability. However, once in the beam, the copy mode quickly takes over.
This explains why large beam settings in Figure~\ref{fig:copy_synth_beam} are more susceptible to copy noise compared to smaller settings.
Thus, while larger beam widths are effective in finding higher likelihood outputs, such sequences may correspond to copies of the source sentence, which explains the drop in BLEU score for larger beams. Deteriorating accuracy of larger beams has been previously observed~\citep{koehn17}, however, it has not until now been linked to the presence of copies in the training data or model outputs.

Note that this finding does not necessarily imply a failure of beam nor a failure of the model to match the data distribution. Larger beams do find more likely hypotheses. It could very well be that the true data distribution is such that no good translation individually get a probability higher than the rate of copy. In that case, even a model perfectly matching the data distribution will return a copy of the source. We refer the reader to \textsection\ref{subsec:seq_level_calibration} for further analysis on this subject. The only conclusion thus far is that \textbf{extrinsic uncertainty is (at least partially) responsible for the degradation of performance of large beams}.

\begin{figure}[t]
\centering
\vspace{-.25cm}
\includegraphics[width=\linewidth]{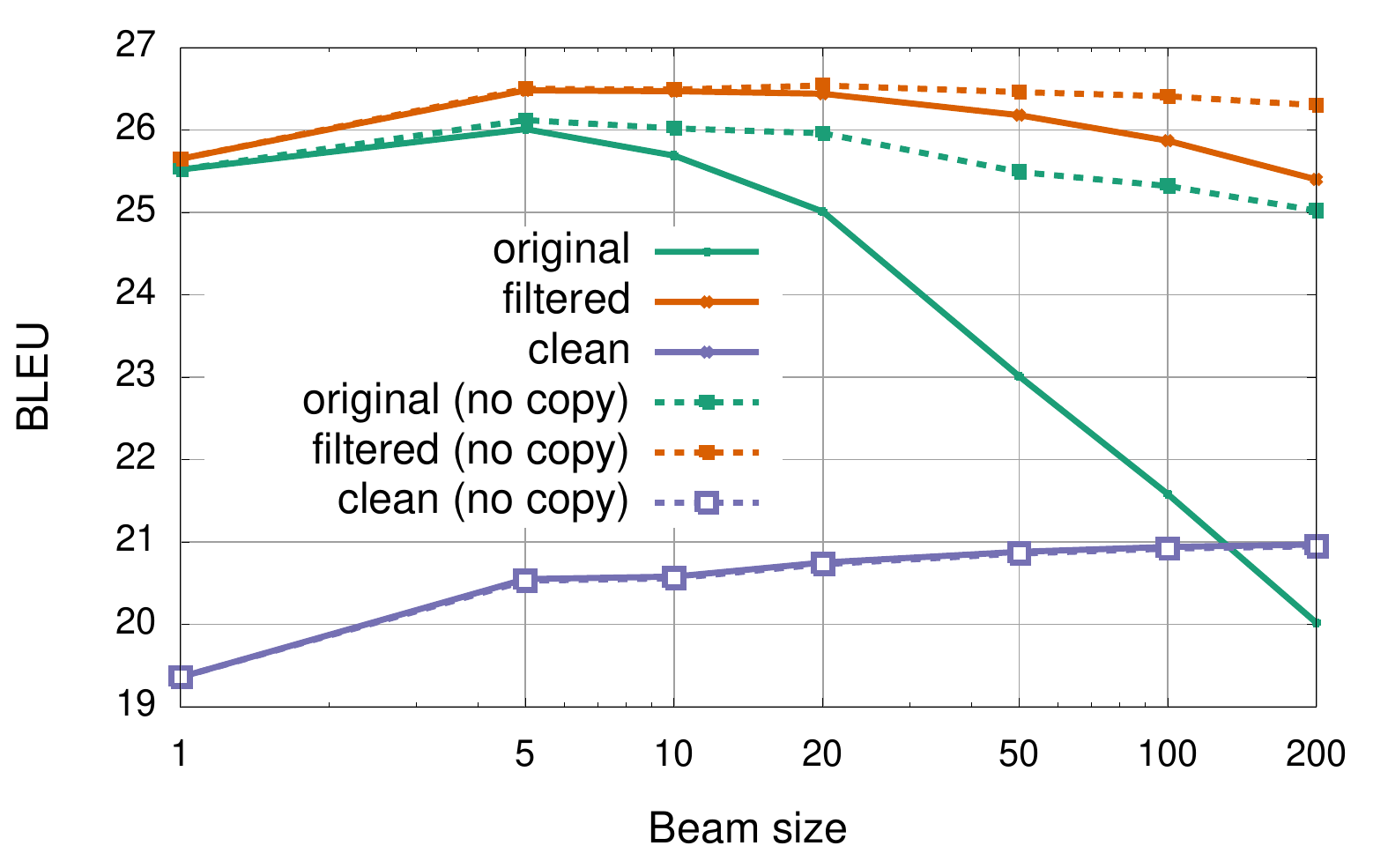}
\vspace{-.75cm}
\caption{\small BLEU on newstest2017 as a function of beam width for models trained on all of the WMT'17 En-De training data (\emph{original}), a filtered version of the training data (\emph{filtered}) and a small but clean subset of the training data (\emph{clean}). We also show results when excluding copies as a post-processing step (\emph{no copy}).}
\label{fig:copyfilter}
\end{figure}

Finally, we present two simple methods to mitigate this issue. First, we pre-process the training data by removing low scoring sentence-pairs according to a model trained on the news-commentary portion of the WMT'17 English-German data (\emph{filtered}; \textsection\ref{sec:datasets}).
Second, we apply an inference constraint that prunes completed beam search hypotheses which overlap by 50\% or more with the source (\emph{no copy}).
Figure~\ref{fig:copyfilter} shows that BLEU improves as beam gets wider on the clean portion of the dataset. Also, the performance degradation is greatly mitigated by both filtering the data and by constraining inference, with the best result obtained by combining both techniques, yielding an overall improvement of 0.5 BLEU over the \emph{original} model.
Appendix~\ref{app:copynoise} describes how we first discovered the copy noise issue.

% \section{Do NMT models fit data distribution?}
\section{Model Fitting and Uncertainty}
\label{sec:modelVSdata}

The previous section analyzed the most likely hypotheses according to the model distribution.
This section takes a more holistic view and compares the estimated distribution to the true data distribution.
Since exact comparison is intractable and we can only have access to few samples from the data distribution, we propose several necessary conditions for the two distributions to match.
First, we inspect the match for unigram statistics. Second, we move to analyze calibration at the set level and design control experiments to assess probability estimates of sentences. Finally, we compare in various ways samples from the model with human references. We find uncontroversial evidence that the model spreads too much probability mass in the hypothesis space compared to the data distribution, often under-estimating the actual probability of individual hypothesis. Appendix~\ref{app:matchfulldistr} outlines another condition.
\subsection{Matching Conditions at the Token Level}
\begin{figure}[t]
\centering
\vspace{-.25cm}
  \includegraphics[width=0.95\linewidth]{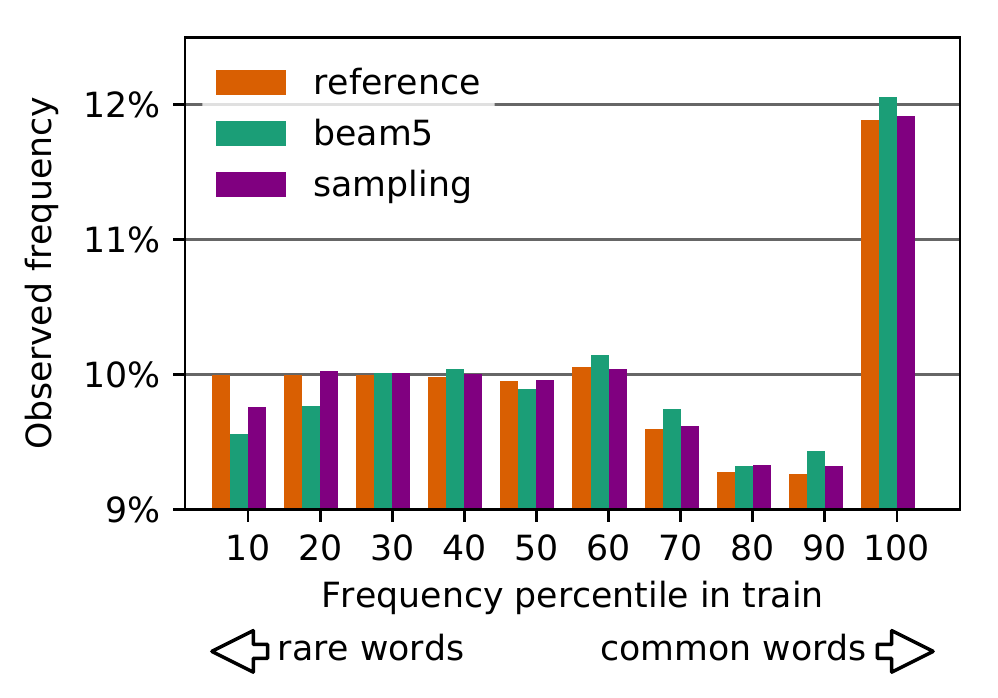}
  \vspace{-.5cm}
\caption{\small Unigram word frequency over the human references, the output of beam search ($k=5$) and sampling on a random subset of 300K sentences from the WMT'14 En-Fr training set.}
\label{fig:unigram_freq}
\end{figure}
If the model and the data distribution match, then unigram statistics of samples drawn from the two distributions should also match (not necessarily vice versa).
This is a particularly interesting condition to check since NMT models are well known to under-estimate rare words~\citep{koehn17};
is the actual model poorly estimating word frequencies or is this just an artifact of beam search?
Figure~\ref{fig:unigram_freq} shows that samples from the model have roughly a similar word frequency distribution as references in the training data,
except for extremely rare words
%; note that in the 10 percent quantile bin, the median occurrence of words is only 12
(see Appendix~\ref{app:moredata} for more analysis of this issue).
On the other hand, beam search over-represents frequent words and under-represents more rare words, 
which is expected since high probability sequences should contain more frequent words.

\begin{figure}[t]
\centering
\vspace{-.25cm}
\includegraphics[width=\linewidth]{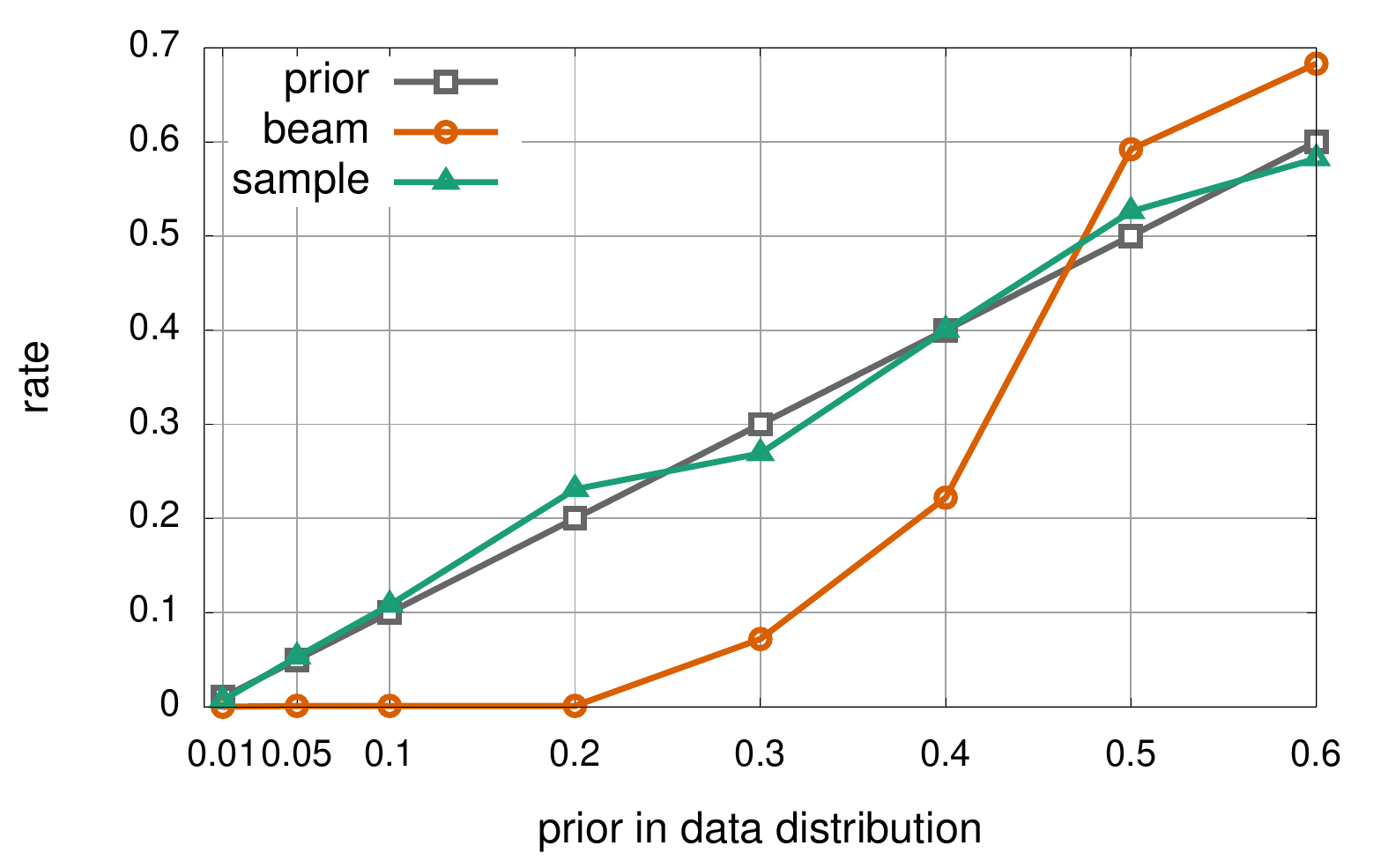}
\vspace{-.75cm}
\caption{\small Comparison of how often a word type is output by the model with beam search or sampling compared to the data distribution; prior is the data distribution. Values below prior underestimate the data distribution and vice versa.} % values above overestimate it.}
\label{fig:replace_synth}
\end{figure}
Digging deeper, we perform a synthetic experiment where we select 10 target word types $w \in W$ and replace each $w$ in the training set with either $w_1$ or $w_2$ at a given \emph{replacement rate} $p(w_1|w)$.\footnote{Each replaced type has a token count between 3k-7k, corresponding to bin 20 in Fig.~\ref{fig:unigram_freq}. $|W| =$ 50k.}
We train a new model on this modified data and verify whether the model can estimate the original replacement rate that determines the frequency of $w_1$ and $w_2$.
Figure~\ref{fig:replace_synth} compares the replacement rate in the data (prior) to the rate measured over the output of either beam search or sampling. Sampling closely matches the data distribution for all replacement rates but beam greatly overestimates the majority class: it either falls below the prior for rates of 0.5 or less, or exceeds the prior for rates larger than 0.5. These observations confirm that the \textbf{model closely matches unigram statistics except for very rare words, while beam prefers common alternatives to rarer ones}.

\subsection{Matching Conditions at the Sequence Level}
\label{subsec:seq_level_calibration}
In this section, we further analyze how well the model captures uncertainty in the data distribution via a sequence of necessary conditions operating at the sequence level.

\textbf{Set-Level Calibration.}
Calibration~\citep{guo17,structpredcalib} verifies whether the model probability estimates $p_m$ match the true data probabilities $p_d$.
If $p_d$ and $p_m$ match, then for any set $S$, we observe:
\[
\mathop{\mathbb{E}}_{x \sim p_d}[\mathbb{I}\{x \in S\}]
%\stackrel{?}{=} 
= p_m(S).
\]
The left hand side gives the expected rate at which samples from the data distribution appear in $S$; the right hand side sums the model probability estimates over $S$.

\begin{figure}[t]
\centering
\vspace{-.25cm}
\includegraphics[width=0.8\linewidth]{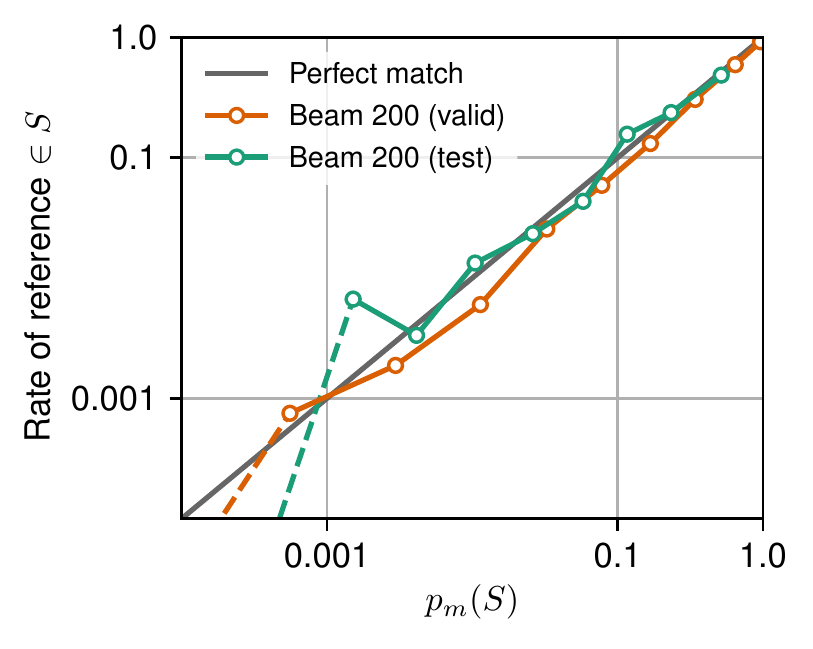}
%\vspace{-.5cm}
\caption{\small Matching distributions at the set level using 200 beam search hypotheses on the WMT'14 En-Fr valid and test set.
Points are binned so that each represents 10\% of sentences. The lowest probability bin (not shown) has value 0 (reference never in $S$).
}
\label{fig:calibration_exp}
\end{figure}

In Figure~\ref{fig:calibration_exp}, we plot the left hand side against the right hand side where $S$ is a set
of $200$ beam search hypotheses on the WMT'14 En-Fr validation set, covering an average of 22.4\% of the model's probability mass.
Points are binned so that each point represents 10\% of sentences in the validation or test set~\citep{nguyen2015emnlp}. For instance, the rightmost point in the figure corresponds to sentences for which beam collects nearly the entire probability mass, typically very short sentences.
This experiment shows that the \textbf{model matches the data distribution remarkably well at the set level} on both the validation and test set.

\begin{figure}[t]
\centering
\vspace{-.25cm}
\includegraphics[width=0.9\linewidth]{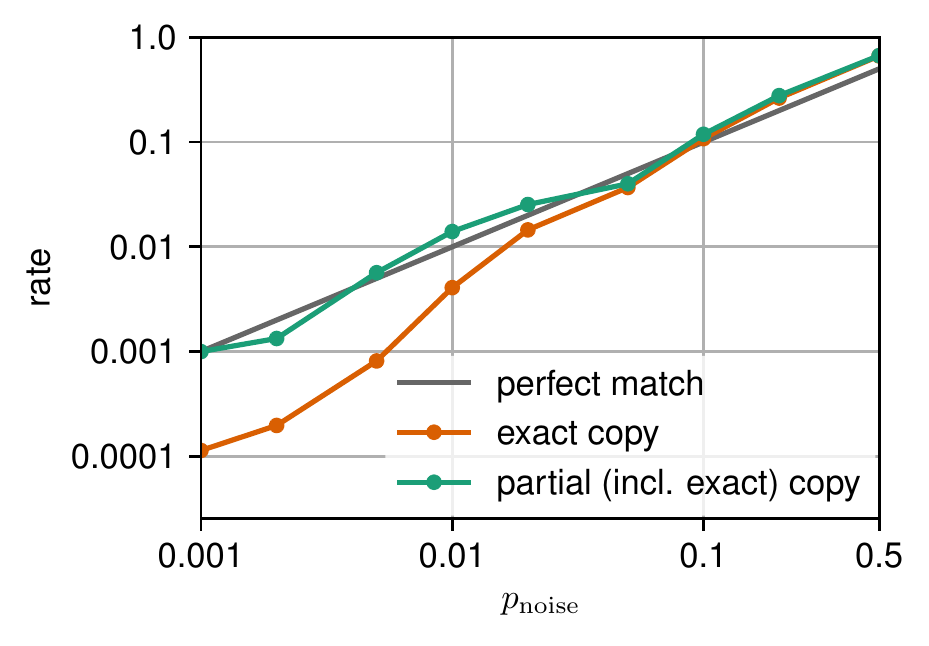}
\vspace{-.5cm}
\caption{\small 
Rate of copy of the source sentence (exact and partial) as a function of the amount of copy noise present in the model's train data (\textsection\ref{sec:beam_failure}). Results on WMT'17 En-De validation set.
}
\label{fig:copy_synth_seqprob}
\end{figure}
\textbf{Control Experiment.}
To assess the fit to the data distribution further, we re-consider the models trained with varying levels of copy noise
($p_{\rm noise}$, cf. \textsection\ref{sec:beam_failure}) and check if we reproduce the correct amount of copying (evaluated at the sequence level)
when sampling from the model.
Figure~\ref{fig:copy_synth_seqprob} shows a large discrepancy: at low $p_{\rm noise}$ the model underestimates the probability of copying (i.e., too few of the produced samples are \emph{exact copies} of the source),
while at high noise levels it overestimates it. 
Moreover, since our model is smooth, it can assign non-negligible probability mass to partial copies\footnote{Partial copies are identified via the IoU at 50\% criterion (\textsection\ref{sec:extrinsic}).}
which are \emph{not} present in the training data.
%We test this by adding partial source copies to the exact source copies, using the IoU at 50\% criterion, cf.~\textsection\ref{sec:extrinsic}.
When we consider both partial and exact copies, the model correctly reproduces the amount of copy noise present in the training data.
%adding partial copies improves the match between the observed rate of copying and $p_{\rm noise}$.
%However, if our model were to perfectly match the data distribution, then it would not assign any probability mass to partial copies since they do not appear in the training data.
%In this case, the lines for ``exact match" and ``partial" would overlap, however, this is clearly not the case.
%However, notice that since the model is smooth,
% it assigns non-negligible probability mass also to partial copies (which are not present at all in the training data).
% In fact, if we include also partial copies (using the same IoU at 50\% criterion of cf. \textsection\ref{sec:extrinsic}), then the model matches more closely the
% rate of copy in the data. 
Therefore, although the model appears to under-estimate some hypotheses at low copy rates, it actually \textbf{smears probability mass in the hypothesis space}. Overall, this is the first concrete evidence of the model distribution not perfectly fitting the data distribution.

\textbf{Expected Inter-Sentence BLEU} is defined as
$$
\mathop{\mathbb{E}}_{x\sim p, x'\sim p}[BLEU(x,x')]
$$
which corresponds to the expected BLEU between two translations sampled from a distribution $p$ where $x$ is the hypothesis and $x'$ is the reference.
If the model matches the data distribution, then the expected BLEU computed with sentences sampled from the model distribution $p_m$ should match
the expected BLEU computed using two independent reference translations (see \textsection\ref{sec:multref} for more details on data collection).

We find that the expected BLEU is $44.5$ and $32.1$ for human translations on the WMT'14 En-Fr and WMT'14 En-De datasets, respectively.\footnote{We also report inter-human pairwise corpus BLEU: 44.8 for En-Fr and 34.0 for En-De; and concatenated corpus BLEU over all human references: 45.4 for En-Fr and 34.4 for En-De.} However,
the expected BLEU of the model is only $28.6$ and $24.2$, respectively. This large discrepancy provides further evidence that
 \textbf{the model spreads too much probability mass across sequences}, compared to what we observe in the actual data distribution.

\subsection{Comparing Multiple Model Outputs to Multiple References} \label{sec:multref}
Next we assess if model outputs are similar to those produced by multiple human translators.
We collect 10 additional reference translations from 10 distinct humans translators for each of 500 sentences randomly selected from the WMT'14 En-Fr and En-De test sets.
We also collect a large set of translations from the model via beam search ($k=200$) or sampling.
We then compute two versions of \emph{oracle BLEU} at the sentence-level:
(i) \textit{oracle reference} reports BLEU for the most likely hypothesis with respect to its best matching reference (according to BLEU);
and (ii) \textit{average oracle} computes BLEU for every hypothesis with respect to its best matching reference and averages this number over all hypotheses.
Oracle reference measures if one of the human translations is similar to the top model prediction, while average oracle indicates
whether most sentences in the set have a good match among the human references.
The average oracle will be low if there are hypotheses that are dissimilar from all human references, suggesting a possible mismatch between the model
and the data distributions.

Table~\ref{tab:oracleBLEU} shows that beam search (besides degradation due to copy noise) produces not only top scoring hypotheses that are very good
(single reference scoring at 41 and oracle reference at 70) but most hypotheses in the beam are close to a reference translation
(as the difference between oracle reference and average oracle is only 5 BLEU points). Unfortunately, beam hypotheses lack diversity and are all
close to a few references as indicated by the coverage number, which measures how many \emph{distinct} human references are matched to at least one of the hypotheses.
In contrast, hypotheses generated by sampling exhibit opposite behavior: the quality of the top scoring hypothesis is lower, several
hypotheses poorly match references (as indicated by the 25 BLEU points gap between oracle reference and average oracle) but coverage is much higher.
This finding is again consistent with the previous observation that the \textbf{model distribution is too spread in hypothesis space}.
We conjecture that the excessive spread may also be partly responsible for the lack of diversity of beam search,
as probability mass is spread across similar variants of the same sequence even in the region of high likelihood. This over-smoothing might be due to the function class of NMT; for instance, it is hard for a smooth class of functions to fit a delta distribution (e.g., a source copy), without spreading probability mass to nearby hypotheses (e.g., partial copies), or to assign exact 0 probability in space, resulting in an overall under-estimation of hypotheses present in the data distribution.

\begin{table}[t]
\small
\centering
\begin{tabular}{lccc}
\toprule
& \multicolumn{2}{c}{\bf beam} & \multicolumn{1}{c}{\bf sampling} \\
& $k=5$ & $k=200$ & $k=200$ \\ % & $k=200$/10k (filter copies) \\
\midrule
\bf Prob.\ covered & 4.7\% & 11.1\% & 6.7\% \\ % & 10.8\% \\
\midrule
\multicolumn{3}{l}{\bf Sentence BLEU} \\
\quad single reference & 41.4 & 36.2 & 38.2 \\ % (39.3) & 32.8 (41.1) \\
\quad oracle reference & 70.2 & 61.0 & 64.1 \\ % (66.3) & 55.4 (69.4) \\
\quad average oracle & 65.7 & 56.4 & 39.1 \\ % (39.2) & 55.4 (56.1) \\
\quad \ -\ \# refs covered & 1.9 & 5.0 & 7.4 \\ % (7.4) & 5.9 (5.7) \\
\midrule
\multicolumn{3}{l}{\bf Corpus BLEU (\texttt{multi-bleu.pl})} \\
\quad single reference & 41.6 & 33.5 & 36.9 \\
\quad 10 references & 81.5 & 65.8 & 72.8 \\
\bottomrule
\end{tabular}
\caption{\small Sentence and corpus BLEU for beam search hypotheses and 200 samples on a 500 sentence subset of the WMT'14 En-Fr test set.
``Single reference" uses the provided reference and the most likely hypothesis,
while oracle reference and average oracle are computed with 10 human references.\label{tab:oracleBLEU}}
\vspace{-0.25cm}
\end{table}
\normalsize
\section{Conclusions and Final Remarks} \label{sec:conclusions}
In this study we investigate the effects of uncertainty in NMT model fitting and search.
We found that search works remarkably well. While the model is generally well calibrated both at the token and sentence level, it tends to diffuse probability mass too much. We have not investigated the causes of this, although we
surmise that it is largely due to the class of smooth functions that NMT models can represent. We instead investigated some of the effects of this mismatch. In particular, excessive probability spread causes poor quality samples from the model. It may also cause the ``copy mode'' to become more prominent once the probability of genuine hypotheses gets lowered. We show that
this latter issue is linked to a form of extrinsic uncertainty which causes deteriorating accuracy with larger beams. Future work will investigate even better tools to analyze distributions and leverage this analysis to design better models.

%Modern translation models capture some uncertainty in the data distribution.
%For instance, large beams return hypotheses that align to several distinct human translation, and model samples show similar word frequencies as observed in the data distribution.
%However, we find that the model distribution does not perfectly match the data distribution at the sequence-level.
%For example,
%we find that random pairs of model hypotheses tend to be much more dissimilar than pairs of human translations, suggesting that
%the model distribution is more diffuse than the data distribution.
%We also analyze the impact of uncertainty on beam search. Small beams are remarkably efficient in approximating the maximum-a-posteriori of the distribution since sampling 10k hypotheses from the model often fails to produce more likely hypotheses than a beam of size $5$.
%We identify copy noise, a form of extrinsic uncertainty, as the underlying cause of deteriorating accuracy with larger beams and propose simple and effective solutions to the problem.
%Greedy search and small beams favor sentences with universally high scores, which can filter out spurious modes.

%Overall, our study suggests that NMT models are good at capturing uncertainty in the head of the distribution, but diffuse the probability mass in the tail over more sentences.
%We hope that these findings will help to design better models and we suggest future work to focus on metrics that better account for uncertainty.

\section*{Acknowledgements}

We thank the reviewers, colleagues at FAIR and Mitchell Stern for their helpful comments and feedback.

%\clearpage
\bibliography{naaclhlt2018}
\bibliographystyle{icml2018}

% \end{document}

\clearpage
\appendix
\appendixpage
%\section{Supplementary Material}
In this supplementary material, we report three additional findings. 
First, we discuss the experiments that made us realize that larger beam degradation was due to copy noise in the training data. 
Second, we introduce another necessary condition for the model and the data distribution to match, which is based on the observation that for some source sentences we do have access to hundreds of references, and therefore, we can directly check whether the two distributions match over the set of unique references.
And finally, we provide a more in depth analysis of the unigram statistics mis-match.

\section{How We Discovered Copy Noise}
\label{app:copynoise}
%\begin{figure}[!b]
%\centering
%\includegraphics[width=\linewidth]{scatterplot_arrow.pdf}
%\caption{Scatter plot showing log-probability and BLEU of samples drawn from the model for four sentences taken from the test set of WMT'14 En-Fr (each color corresponds to a different test  sentence). (1) shows samples where the model copied the source sentence, yielding very large likelihood but low BLEU. (2) and (3) are valid translations of the same source sentence, except that (2) is a cluster of samples using different choice of words.}
%\label{fig:scatterplot_arrow}
%\end{figure}
In this section we report the initial experiment which led us to the realization that degradation of large beams is due to noise in the training data, as the process may
be instructive also for other researchers working in this area.

\begin{figure}[t]
\centering
\includegraphics[width=\linewidth]{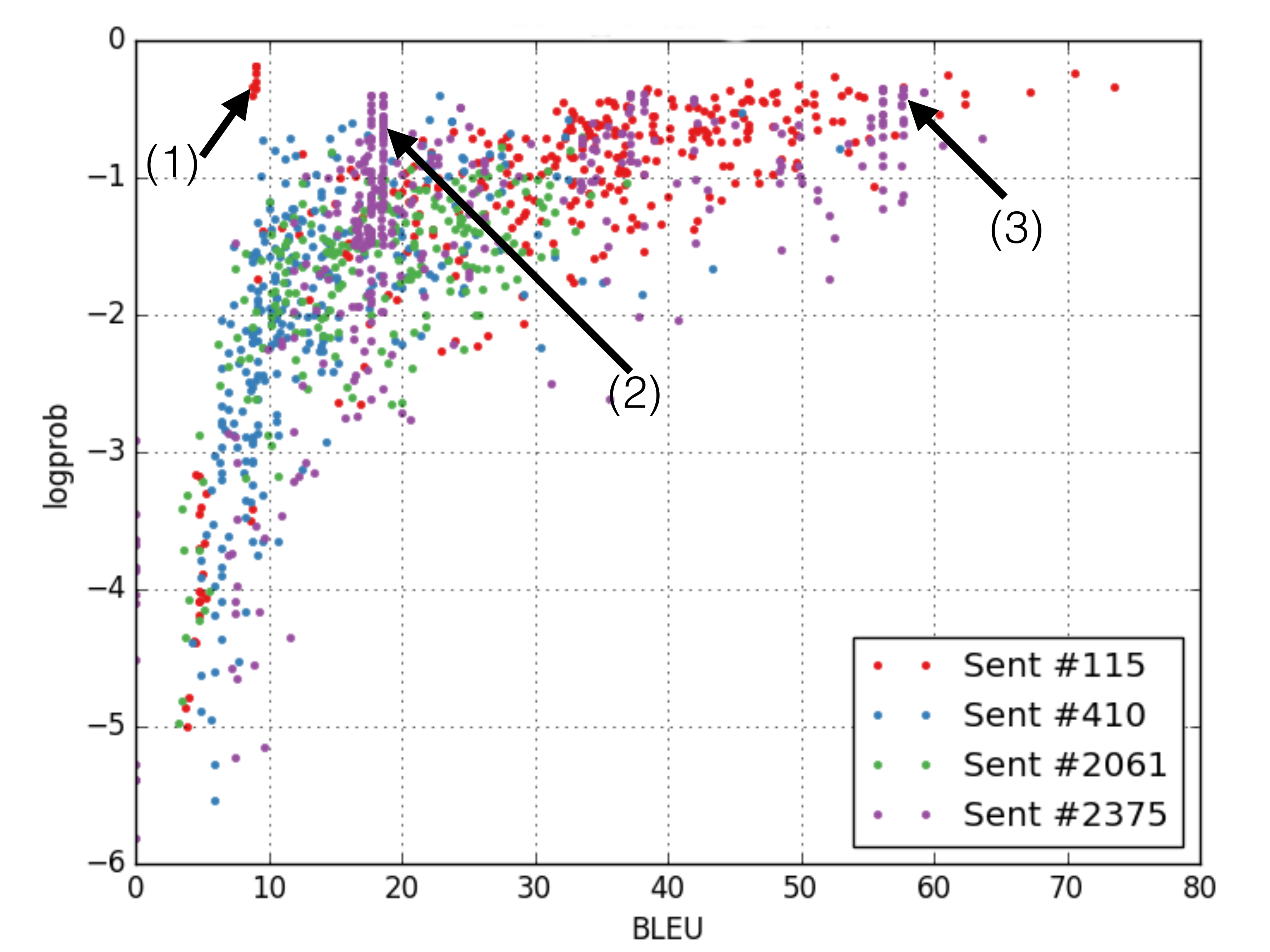}
\caption{\small Scatter plot showing log-probability and BLEU of samples drawn from the model for four sentences taken from the test set of WMT'14 En-Fr (each color corresponds to a different test  sentence). (1) shows samples where the model copied the source sentence, yielding very large likelihood but low BLEU. (2) and (3) are valid translations of the same source sentence, except that (2) is a cluster of samples using different choice of words.}
\label{fig:scatterplot_arrow}
\end{figure}

A nice visualization of samples drawn from the model is via a scatter plot of log-probability VS.\ BLEU, as shown in Figure~\ref{fig:scatterplot_arrow} for four sentences picked at random from the test set of WMT'14 En-Fr.

First, this plot shows that while high BLEU implies high log-likelihood, the vice versa is not true, as low BLEU scoring samples can have wildly varying log-likelihood values.

Second, the plot makes very apparent  that there are some outlier hypotheses that nicely cluster together.

For instance, there are two clusters corresponding to sentence id 2375, marked with (2) and (3) in Figure~\ref{fig:scatterplot_arrow}. These clusters have relatively high log-likelihood but very different BLEU score. The source sentence is:\\
``\textit{Should this election be decided two
months after we stopped voting?}''\\
The target reference is:\\
``\textit{Cette \'election devrait-elle \"etre
d\'ecid\'e deux mois apr\`es que le vote est termin\'e?}''\\
while a sample from cluster (2) is:\\
``\textit{Ce choix devrait-il \"etre d\'ecid\'e deux
mois apr\`es la fin du vote?}''\\
and a sample from cluster (3) is:\\
``\textit{Cette \'election devrait-elle \"etre
d\'ecid\'ee deux mois apr\"es l'arr\"et du scrutin?}''\\
This example shows that translation (2), which is a valid translation, gets a low BLEU because of a choice of a synonym word with different gender which causes all subsequent words to be inflected differently, yielding overall a very low n-gram overlap with the reference, and hence a low BLEU score. This is an example of the model nicely capturing (intrinsic) uncertainty, but the metric failing at acknowledging that.

Let's now look at cluster (1) of sentence id 115. This cluster achieves extremely high log-likelihood but also extremely low BLEU score. The source sentence is:\\
``\textit{The first nine episodes of Sheriff [unk]'s Wild West will be available from November 24 on the site [unk] or via its application for mobile phones and tablets.}''\\
The target reference is:\\
``\textit{Les neuf premiers \'episodes de [unk] [unk] s Wild West seront disponibles \`a partir du 24 novembre sur le site [unk] ou via son application pour t\'el\'ephones et tablettes.}''\\
while a sample from cluster (1) is:\\
``\textit{The first nine episodes of Sheriff [unk] s Wild West will be available from November 24 on the site [unk] or via its application for mobile phones and tablets.}''\\
In this case, the model copies almost perfectly the source sentence. Examples like these made us discover the ``copy issue", and led us to then link beam search degradation to systematic mistakes in the data collection process.

In conclusion, lots of artifacts and translation issues can be easily spotted by visualizing the data and looking at clusters of outliers.

\newpage
\section{Another Necessary Condition: Matching the Full distribution for a Given Source}
\label{app:matchfulldistr}
\begin{figure}[t]
\centering
\includegraphics[width=\linewidth]{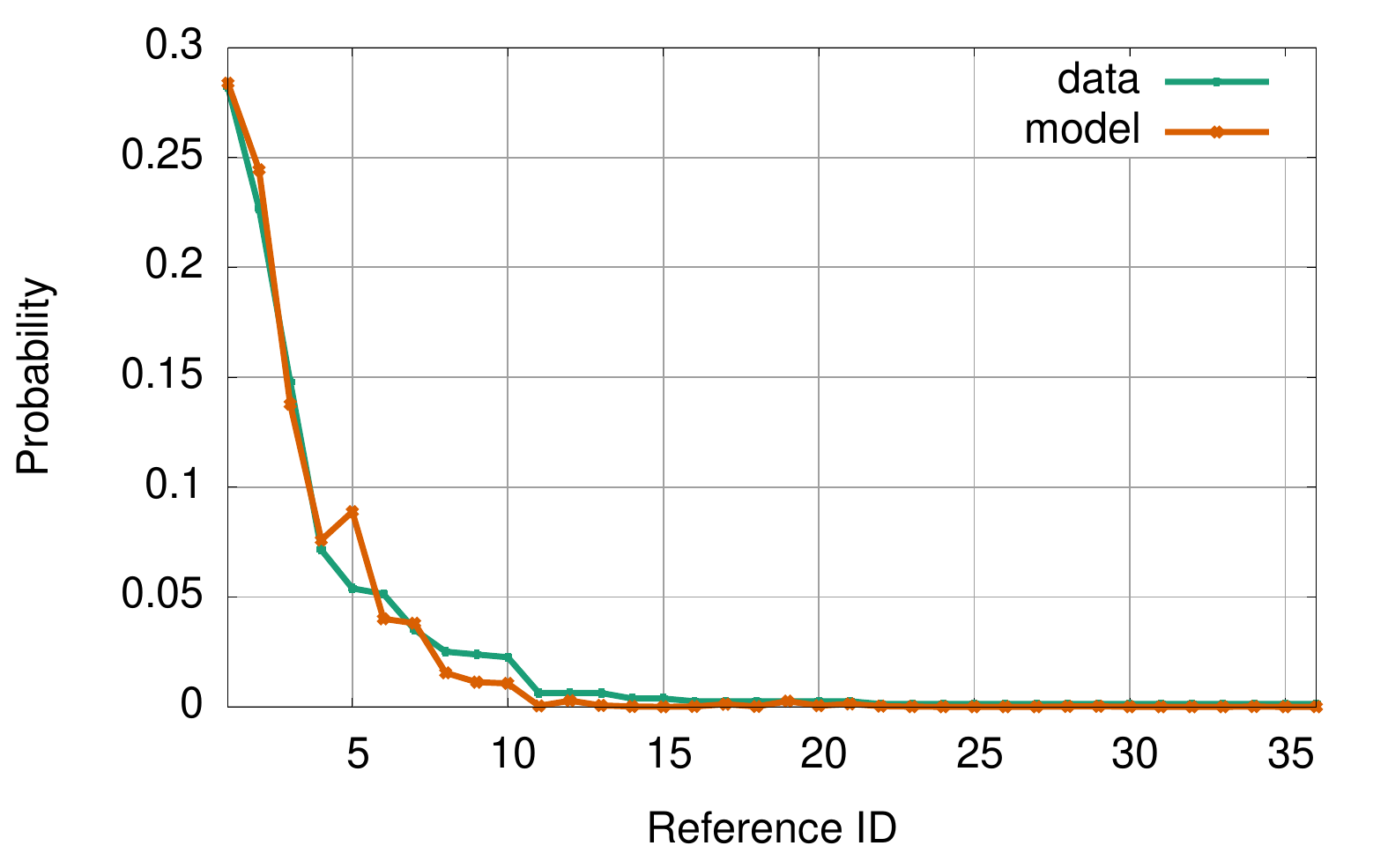}
\caption{\small Comparison between the data and the model distributions for the source sentence ``(The president cutoff the speaker)''. The data distribution is estimated over 798 references of which 36 are unique. The hypotheses of the data distribution (x-axis) are sorted in descending order of empirical probability mass. The model matches rather well the data distribution.}
\label{fig:probabDataModelAnecdotal}
\end{figure}

In \textsection\ref{sec:modelVSdata} we have investigated several necessary conditions for the model distribution to match the data distribution. Those conditions give an aggregate view of the match and they are mostly variants of calibration techniques, whereby the data distribution is approximated via Monte Carlo samples (human translations), since that is all we have access to.

Ideally, we would like to check the two distributions by evaluating their mass at every possible target sequence, but this is clearly intractable and not even possible since we do not have access to the actual data distribution.

However, there are sentences in the training set of WMT'14 En-Fr (EuroParl corpus) that appear several times. For instance, the source sentence ``(The President cut off the speaker)." appears almost 800 times in the training set with 36 unique translations. For such cases, we can then have an accurate estimate of the ground truth data distribution (for that given source sentence) and check the match with the model distribution. This is yet another necessary condition: if the model and data distribution match, they also match for a particular source sentence.

Figure~\ref{fig:probabDataModelAnecdotal} shows that for this particular sentence the model output distribution closely matches the data distribution.

\newpage
\section{Does More Data Help?}
\label{app:moredata}
\begin{figure}[!t]
\centering
  \includegraphics[width=0.95\linewidth]{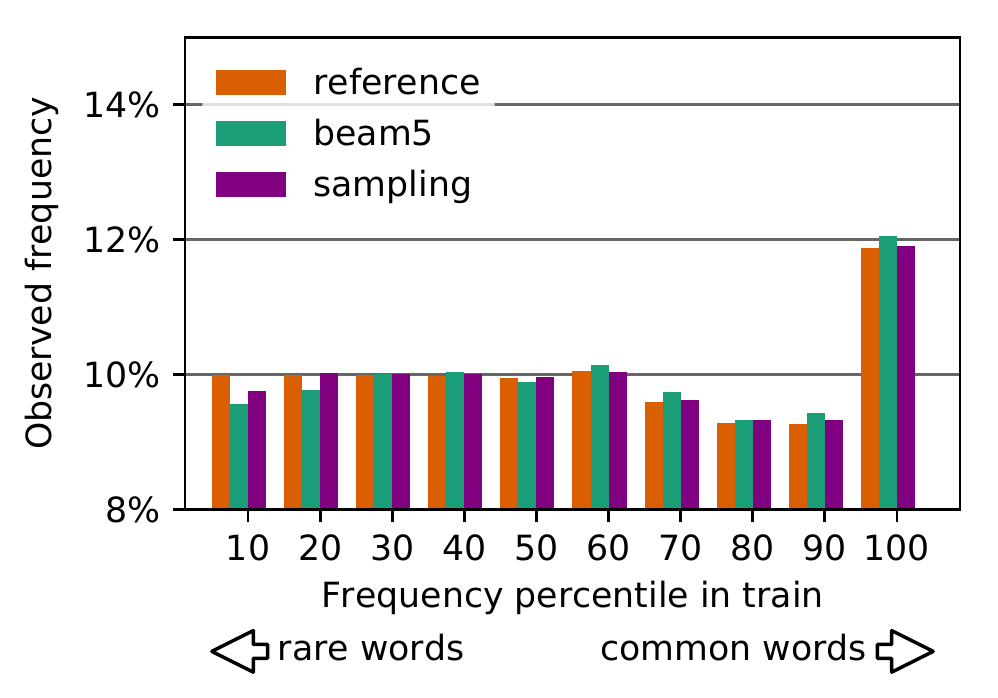}
  \includegraphics[width=0.95\linewidth]{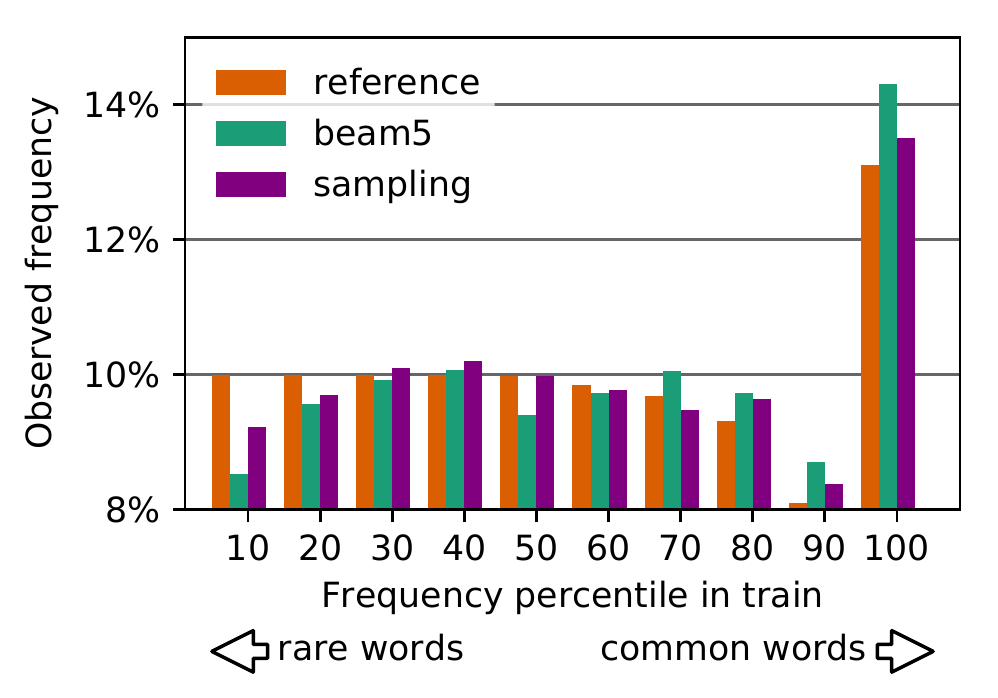}
\caption{\small Unigram word frequency over the human references, the output of beam search ($k=5$) and sampling on the WMT'14 En-Fr (top) and WMT'17 En-De news-commentary portion (bottom) of the training set.}
\label{fig:unigram_freq_comparison}
\end{figure}
The findings reported in this paper are quite robust to the choice of architecture as well as dataset.
For instance, we compare in Figure~\ref{fig:unigram_freq_comparison} 
the binned unigram word frequencies on the smaller news-commentary portion of the WMT'17 En-De dataset with the larger WMT'14 En-Fr dataset (which was already reported in Figure~\ref{fig:unigram_freq}). 
The En-Fr data is about 100 times bigger than the En-De news-commentary dataset, as described in \textsection\ref{sec:datasets} and the En-Fr model performs much better than the En-De model, with
a BLEU of 41 versus only 21 (see Table~\ref{tab:humanstudy} and Figure~\ref{fig:copyfilter}). 
%Despite the better quality of the En-Fr model,
We observe the same tendency of the model to under-estimate very rare words 
(compare \texttt{beam5} vs.~\texttt{reference} in the 10 percentile bin). 
However, the under-estimation is much more severe in the En-De model, nearly 1.5\% as opposed to only 0.4\%.
Note that the median frequency of words in the 10 percentile bin is only 12 for the En-De dataset, but is 2552 for the En-Fr dataset.
The NMT model clearly needs more data to better estimate its parameters and fit the data distribution.

%\newpage
%\section{Do Lengths Match?}
%
%\begin{figure}[t]
%\centering
%\includegraphics[width=\linewidth]{samples_lengths}
%\caption{\small Average length of the highest-probability sample as we increase the number of samples that we draw, on the WMT'14 En-Fr validation set. The average length of the reference translation and top-scoring beam hypothesis ($k = 200$) are shown for comparison.}
%\label{fig:lengthmismatch}
%\end{figure}
%
%In \textsection\ref{sec:beamisgood} we observed that BLEU and model probability are imperfectly correlated.
%In particular, while we can draw more samples from the model and find increasingly likely outputs, these outputs do not necessarily receive larger BLEU scores.
%In fact, the average BLEU score of the highest-probability sample decreases after drawing just 297 samples (cf. Figure~\ref{fig:oracle_bleu}, \emph{Right}).
%
%We found that ``copying" is one cause of this deterioration, but another cause is due to length mismatch.
%In Figure~\ref{fig:lengthmismatch} we plot the average length of the highest-probability sample as a function of the number of samples that we draw.
%Note that we choose the highest-probability sample based on the average per-token probability, thus it is normalized by length.
%Nevertheless, we observe that sampled outputs are shorter, on average, than either the reference or top beam hypothesis.
%Moreover, as we draw increasingly likely samples, the average length of the sampled outputs decreases.
%We speculate that this is because 

\end{document}